\documentclass[runningheads]{llncs}
\usepackage{graphicx}

\usepackage{tikz}
\usepackage{comment} 
\usepackage{amsmath,amssymb} 
\usepackage{color}

\usepackage{subfigure}
\usepackage{wrapfig}
\usepackage{array}
\usepackage{multirow}
\usepackage{booktabs}
\usepackage{subfigure}
\def\ie{\emph{i.e.}}

\usepackage[breaklinks=true,letterpaper=true,colorlinks,bookmarks=false]{hyperref}

\begin{document}
\pagestyle{headings}
\mainmatter
\def\ECCVSubNumber{100}  

\title{AutoTrajectory: Label-free Trajectory Extraction and Prediction from Videos using Dynamic Points} 

\titlerunning{AutoTrajectory: Label-free Trajectory Extraction and Prediction}
%
\author{Yuexin Ma$^{\star}$\inst{1} \and
Xinge Zhu$^{\star}$\inst{2} \and
Xinjing Cheng\inst{3} \\ Ruigang Yang\inst{3} \and Jiming Liu\inst{1} \and Dinesh Manocha\inst{4}}
\authorrunning{Y. MA et al.}
%
\institute{$^1$ Hong Kong Baptist University $^2$ Chinese University of Hong Kong \\ $^3$ Inceptio $^4$ University of Maryland at College Park}
\maketitle

\begin{abstract}
Current methods for trajectory prediction operate in supervised manners, and therefore require vast quantities of corresponding ground truth data for training. In this paper, we present a novel, label-free algorithm, AutoTrajectory, for trajectory extraction and prediction to use raw videos directly. To better capture the moving objects in videos, we introduce dynamic points. We use them to model dynamic motions by using a forward-backward extractor to keep temporal consistency and using image reconstruction to keep spatial consistency in an unsupervised manner. Then we aggregate dynamic points to instance points, which stand for moving objects such as pedestrians in videos. Finally, we extract trajectories by matching instance points for prediction training. To the best of our knowledge, our method is the first to achieve unsupervised learning of trajectory extraction and prediction. We evaluate the performance on well-known trajectory datasets and show that our method is effective for real-world videos and can use raw videos to further improve the performance of existing models.
\end{abstract}

\footnote{\noindent $^{\star}$Equal contribution}

\vspace{-9ex}
\section{Introduction}
For intelligent agents like robots and autonomous vehicles, it is crucial to be able to forecast neighboring traffic-agents' future trajectories for navigation and planning applications. Trajectory prediction for dynamic objects has been widely studied and is an active area of research. Some traditional methods for trajectory prediction are based on motion models such as Bayesian networks~\cite{Lefevre2011ExploitingMI}, Kalman filters~\cite{Baar2001ANA}, Gaussian process regression models~\cite{Hall2003CorrelationbasedFS}, etc. These methods can deal with simple scenarios with very few moving instances, but are limited in complex real-world scenarios with many instances or agents interacting with each other. Recurrent Neural Network (RNN) and its variant long short-term Memory (LSTM) have become an effective way for trajectory prediction due to its ability to model non-linear temporal dependencies in sequence learning and generation~\cite{Palaz2016TowardsES,Cao2015LookAT}. Based on these networks, recent works are able to achieve good accuracy on predictig trajectories for pedestrians~\cite{Alahi2016SocialLH,Gupta2018SocialGS}, vehicles~\cite{Lee2017DESIREDF,Park2018SequencetoSequencePO}, and heterogeneous traffic-agents~\cite{ma2019trafficpredict}. However, all of the above methods operate in supervised manners, which rely heavily on labeled trajectory data. One general method to get a trajectory dataset~\cite{Pellegrini2009YoullNW,Lerner2007CrowdsBE} is to label consecutive positions of moving traffic-agents (pedestrians or vehicles) on fixed-view videos and then transfer the trajectory from the image coordinate system to a real-world coordinate system. Labeling consecutive objects from videos is complex and expensive~\cite{He2018TrackingBA}. There is a great demand for an unsupervised learning method to alleviate the dependence on annotations by simply taking raw videos as input and automatically extracting trajectories for training prediction network. 

The most pivotal and challenging task for label-free trajectory extraction is capturing the moving objects, which we also call dynamic instances, from videos without any supervision. There are some related problems that also need to learn the motion dynamics of objects from videos, like activity prediction~\cite{Luo2017UnsupervisedLO}, video prediction~\cite{Kim2019UnsupervisedKL,Minderer2019UnsupervisedLO}, and object tracking~\cite{He2018TrackingBA,Crawford2019ExploitingSI}. However, we found they did not perform well for common bird's-eye view videos (sometimes, we may just see the heads and shoulders of pedestrians). For such videos, it is difficult to distinguish instances for the network just by appearance and structure features, while the above methods all rely on these features. To extract trajectories for dynamic instances in videos, we need consider not only the appearance and structure features in spatial space, but also the dynamic features (consecutive motions of objects) in temporal space. Our work is based on this consideration.

\textbf{Main Results:} In this paper, we propose a label-free learning-based method AutoTrajectory for trajectory extraction and prediction to overcome the above difficulties. To better capture the motion dynamics of moving objects in the video, we use the concept of \emph{dynamic points}, which can focus on dynamic locations on images. These points are derived by keeping the spatial appearance and structure consistent via self-image reconstruction and maintaining the temporal dynamic features to be consistent in consecutive frames.
Because our target is to get trajectories of instances, then we use optical flow and clustering algorithms to aggregate dynamic points to instances and extract trajectories by the matching method. Finally, we use these trajectories to train the trajectory prediction network. The whole process uses no labels. Our approach contains four main parts, including dynamic point modeling, dynamic-to-instance aggregation, trajectory extraction, and trajectory prediction. The main contributions of our work are:

\vspace{-1ex}
\begin{itemize}
\item We propose a label-free trajectory extraction and prediction pipeline, which can extract trajectories of dynamic instances from raw videos directly and train a prediction network.
\item We propose a novel forward-backward dynamic-point extractor, which could capture dynamic features in consecutive images.
\item We propose a dynamic-to-instance mechanism, which could aggregate dynamic locations to instances.
\item Our method is effective and has good scalability. With more raw videos, our method can also improve existing methods in a semi-supervised manner.
\end{itemize}

\section{Related Work}

\subsection{Trajectory Prediction}
Classical model-based approaches for trajectory prediction~\cite{Lefevre2011ExploitingMI,Baar2001ANA,Hall2003CorrelationbasedFS,ma2018autorvo} focus on the inherent motion regularities of objects themselves. However, the motion of dynamic objects in the real world is diverse and can be governed by many factors, like neighboring objects' motion states and the environment. These methods are limited in modeling complex scenarios. Recently, RNN and its variant LSTM have achieved great success in modeling sequence prediction tasks~\cite{Palaz2016TowardsES,Cao2015LookAT}. Based on these basic networks, many prediction approaches have outperformed classical methods in real-world benchmarks~\cite{Alahi2016SocialLH,Gupta2018SocialGS,Lee2017DESIREDF,Park2018SequencetoSequencePO,ma2019trafficpredict,Chandra2019ForecastingTA,Chandra2019RobustTPET,Tang2019MultipleFP,Sadeghian2018SoPhieAA,Chai2019MultiPathMP,Rhinehart2019PRECOGPC,Zhao2019MultiAgentTF,Pan2019LaneAP,Wang2019UnsupervisedPT}. However, these supervised methods require large-scale, well-annotated trajectory data. Two main ways to generate the data are labeling moving instances from fixed-view videos and LiDAR point clouds. Both ways are expensive and time-consuming. Even though there are a lot of videos captured by street or commodity cameras, they cannot be used to improve the prediction performance without annotation. We try to solve this problem by using unsupervised manners.

\subsection{Supervised Multi-object Tracking}
Except for manual labeling, another possible solution for getting trajectories from videos is using current SOTA trackers. However, most modern trackers follow the tracking-by-detection paradigm~\cite{bewley2016simple,bertinetto2016fully,fang2018recurrent,sharma2018beyond,tang2017multiple,zhu2018online,xu2019spatial}. The performance depends largely on the detector used to find the objects as the tracking targets and the detector requires large-scale labeled data. Besides, the tracker is always trained for fixed-categories, which is hard to adapt to other domains. Recent trend in multi-object tracking is combining both detection and tracking in one framework\cite{feichtenhofer2017detect,ren2015faster,bergmann2019tracking}. However, they do not overcome the above limitations. Our approach focuses on exploring the nature of video, \ie, the dynamic information, which is naturally category-free and works well on all domains.

\subsection{Unsupervised Learning for Dynamic Modeling}
To extract trajectories from sequential frames, a crucial step is learning the motion dynamics of the video. Many works have explored unsupervised methods for dynamic modeling for videos to solve different problems~\cite{Luo2017UnsupervisedLO}. Based on keypoint-based representation~\cite{Jakab2018ConditionalIG}, the video prediction approach~\cite{Minderer2019UnsupervisedLO} could decouple pixel generation from dynamic prediction. ~\cite{Kim2019UnsupervisedKL} combines keypoints and extra action classes to help generate action conditioned image sequences. Inspired by the function of keypoint on video prediction and generation, we designed dynamic point. For unsupervised tracking, unsupervised single object tracking is the mainstream~\cite{untracking2,untracking3,untracking}. However, they cannot handle the scenes with multiple objects. For unsupervised multi-object tracking, the pioneering work AIR~\cite{Eslami2016AttendIR} proposes a VAE-based framework to detect objects from individual images through inference, which is followed by~\cite{Kosiorek2018SequentialAI,He2018TrackingBA}. ~\cite{Crawford2019ExploitingSI} makes use of spatially invariant computations and representations to exploit the structure of objects in videos. In our initial attempts, we applied these unsupervised methods to locate dynamic instances on pedestrian videos directly but got poor results. The primary reason is that the above methods rely on structure and appearance features of objects, which are not applicable for trajectory extraction from bird's-eye view videos, where the these features are not very obvious.

\section{Our Approach}
\subsection{Problem Definition}
Given raw videos without any annotations, our task is to obtain a trajectory predictor in an unsupervised manner. We solve this problem by two main steps: trajectory extraction and trajectory prediction. For trajectory extraction, the input is raw videos captured by street cameras, and the output is $R=\left\{r_1, r_2, ..., r_n\right\}$, where $R$ denotes all trajectories of moving objects in the videos. The trajectory for the $i$th object is defined as a set of discrete positions in the real-world coordinate system: $r_i=\left\{p_i^{t_{start}}, p_i^{t_{start}+1}, ..., p_i^{t_{end}}\right\}$, where $[t_{start}, t_{end}]$ denotes the time interval when the object occurs in the video. For the trajectory prediction, the extracted trajectories $R$ acts as the dataset for training and validating the prediction network. The predictor observes objects' trajectories of an time interval and predicts their trajectories in the following period, like observing trajectory of 3s and predicting the trajectory for the next 5s. Without any label, we finally compute a trajectory prediction predictor.

\subsection{Method Overview}
\begin{figure*}[t]
\begin{center}
   \includegraphics[width=1.0\linewidth]{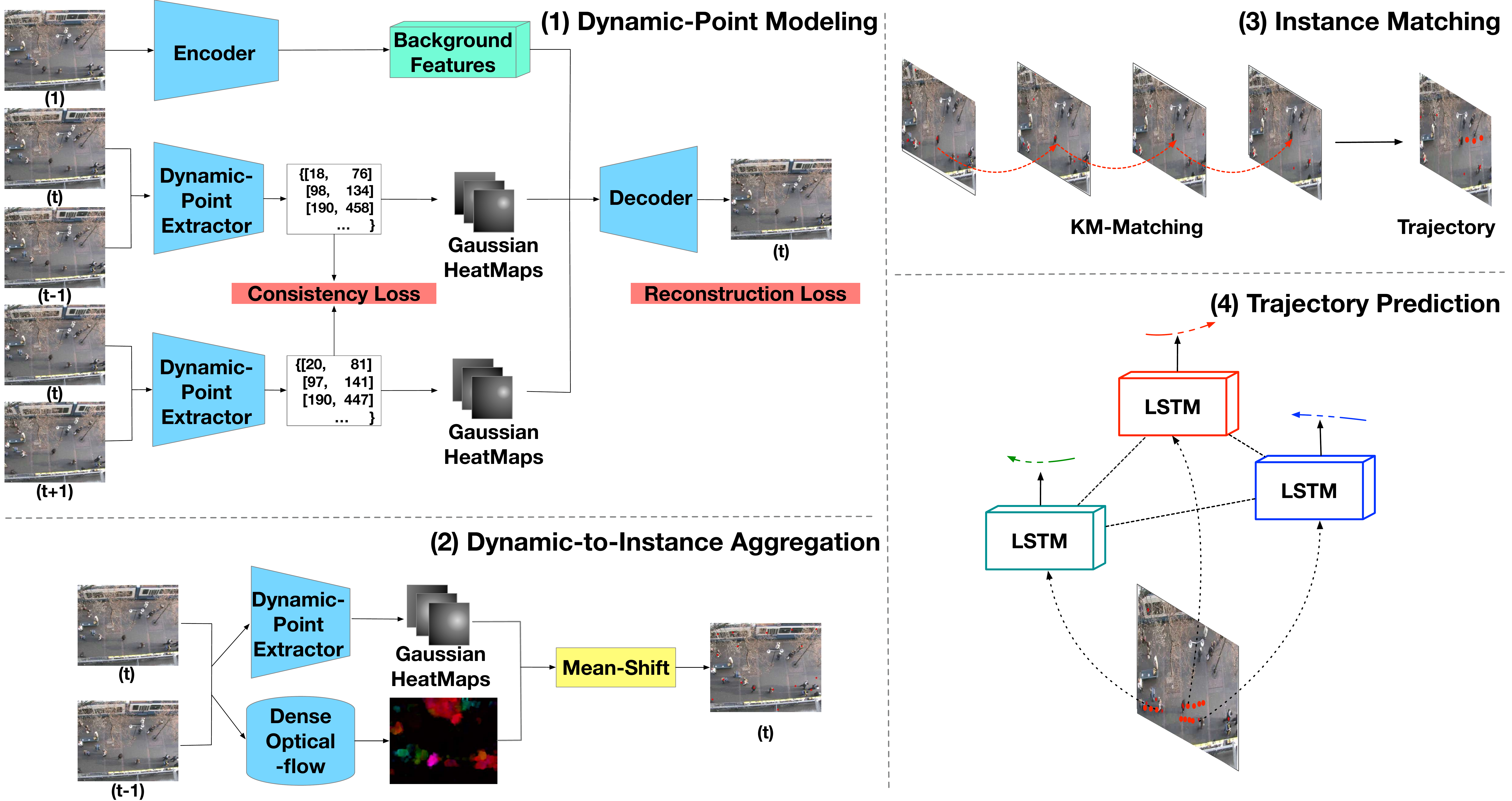}
\end{center}
\vspace{-2ex}
   \caption{Pipeline and main components of AutoTrajectory. Specifically, the first three components form the unsupervised trajectory extraction.}
\label{fig:pileline}
\vspace{-3ex}
\end{figure*}

We propose a label-free pipeline to generate the trajectory and then train the trajectory predictor.
Specifically, our approach consists of four components: Dynamic-Point Modeling, Dynamic-to-Instance Aggregation, Instance Matching, and Trajectory Prediction. The first three parts form the unsupervised trajectory extraction. We show the pipeline in Fig.~\ref{fig:pileline}. In what follows, we will present these components in details.


\subsection{Dynamic-Point Modeling}

This part performs the unsupervised discovery of the dynamic points. Given a sequence of images, including Image$_1$ ($I_1$), Image$_{t-1}$ ($I_{t-1}$), Image$_t$ ($I_t$), and Image$_{t+1}$ ($I_{t+1}$), our objective is to capture $K$ pixel locations, namely dynamic points $\Phi \in \mathbb{R}^{K\times2}$, which correspond to the locations of moving regions in $I_t$.  

The detailed networks are shown in Fig.~\ref{fig:pileline}\textbf{(1)}. The first image provides the background and layout features. Two pairs of consecutive images are used to capture the dynamic points in $I_t$. Both background features and the dynamic-point gaussian heatmaps are used to reconstruct the image ($I_t$). 
The learning objective $\mathcal{L}$ then consists of two parts, consistency loss $\mathcal{L}_C$ and reconstruction loss $\mathcal{L}_R$, to regularize the dynamic points extraction and image reconstruction, respectively. The total objective is formulated as $\mathcal{L} = \mathcal{L}_R + \beta \mathcal{L}_C$.

\noindent \textbf{Forward-Backward Dynamic-Point Extractors.}~~~ Keypoints are known as natural representations of objects. Some methods for video prediction~\cite{Kim2019UnsupervisedKL,Minderer2019UnsupervisedLO} encode single frames to keypoints to make the representation spatially structured and then generate videos. For the trajectory extraction from bird's eye view videos (Fig.~\ref{fig:keyopt}(a)), the movement features in the temporal space are very important due to the limited appearance and structure features. Thus, we extend keypoints to dynamic points by utilizing more consecutive infomation in the temporal space. Dynamic-point extractors use two consecutive images to capture the dynamic points $\Phi$. Two sets of images are applied in both forward (\ie~from $t-1$ to $t$) and backward (\ie~from $t+1$ to $t$) directions to keep the dynamic points of $I_t$ consistent. The consistency loss is a location-wise MSE loss.
\begin{align}
\mathcal{L}_C = ||(\Phi_{forward} - \Phi_{backward})||^2_2.
\end{align}

\noindent \textbf{Gaussian Heatmaps.}~~~ After obtaining dynamic points $\Phi\in \mathbb{R}^{K\times 2}$, we use gaussian heatmaps $\mathcal{H}\in \mathbb{R}^{H\times W \times K}$ to encode these points $\Phi$ into pixel representation, which is more suitable as the input for the convolutional reconstruction network. 
We first normalize the dynamic points via Softmax (\ie~$\Phi^*$ after normalization). Then each dynamic point is replaced with a gaussian function:
\begin{align}
\mathcal{H} = \exp(-\frac{1}{2\sigma^2}\lVert \Phi - \Phi^{*} \rVert^2),
\end{align}
where $\sigma$ is a fixed standard deviation. The result $\mathcal{H}\in \mathbb{R}^{H\times W \times K}$ is the gaussian heatmap that describes the dynamic locations; it is also used as an input to the decoder network.

\noindent \textbf{Decoder.}~~~ The decoder network utilizes background and layout features and dynamic-point heatmaps to reconstruct the image (\ie~the reconstructed image is $I_t^*$). The reconstruction loss is a pixel-wise L2 loss:
\begin{align}
\mathcal{L}_{R} = \lVert (I_t^* - I_t) \rVert_2^2.
\end{align}
 In this way, the objective could induce the representation of dynamic points for reconstructing the specific image in an unsupervised manner. Meanwhile, image reconstruction can make full use of the appearance and structure information in the spatial space, which is a complement to the focus on dynamic motions.

\subsection{Dynamic-to-Instance Aggregation}

\begin{figure*}[t]
\begin{center}
   \includegraphics[width=1.0\linewidth]{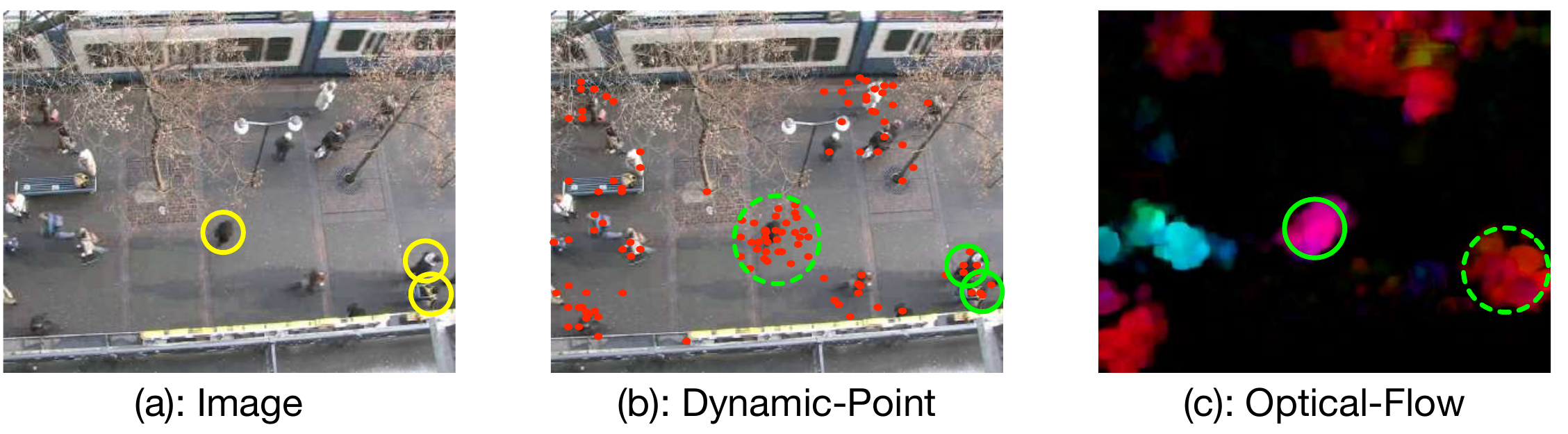}
\end{center}
\vspace{-2ex}
   \caption{A sample of image with dynamic points and optical flow. Yellow circles denote the pedestrians. Green dashed circles denote the poor instance-level representations. Solid green circles indicate the better instance-level descriptions.}
\label{fig:keyopt}
\vspace{-3ex}
\end{figure*}

Dynamic points could detect dynamic locations on images, while trajectories originate from instances. After acquiring the well-trained dynamic-point extractor in the previous step, we aim to group these dynamic points to get the instance-level location information. Intuitively, the solution is to cluster the dynamic points to instance points directly. However, the dynamic points have some characteristics: it shows better instance-level information (distinguishing different objects well)  when multiple objects are close to each other while shows loose when solo object occurs. 

We tackle this problem by introducing the optical flow into the instance-level information collection. An example of the dynamic points and optical flow is shown in Fig.~\ref{fig:keyopt}. We can observe that the dynamic points correspond to a better instance-level representation, when multiple objects are in close proximity (solid green circles in (b)). The optical flow shows the compact representation for solo objects (solid green circle in (c)).
It shows that that dynamic points and optical flow are complementary.  

Specifically, we use a pair of consecutive images ($I_t$ and $I_{t-1}$) to extract the dynamic point representation and optical flow, training the dynamic-point extractor in step 1 and applying unsupervised optical flow method~\cite{Farnebck2003TwoFrameME}. The gaussian heatmaps are upsampled to the original image size via bilinear interpolation. Then both gaussian heatmaps and optical flow are concatenated as the input to the clustering method, \ie~mean-shift, to get the cluster centers, which are the coordinates of objects.

\noindent \textbf{Region of Validity.}~~~ Since there exist invalid regions for moving objects in images (railway in Fig.~\ref{fig:keyopt}) and the background is static for a fixed camera, we apply the region of validity to filter these outliers located in the invalid regions. We show the details in the experiment section.

\subsection{Instance Matching}

\begin{figure*}[t]
\begin{center}
   \includegraphics[width=1.0\linewidth]{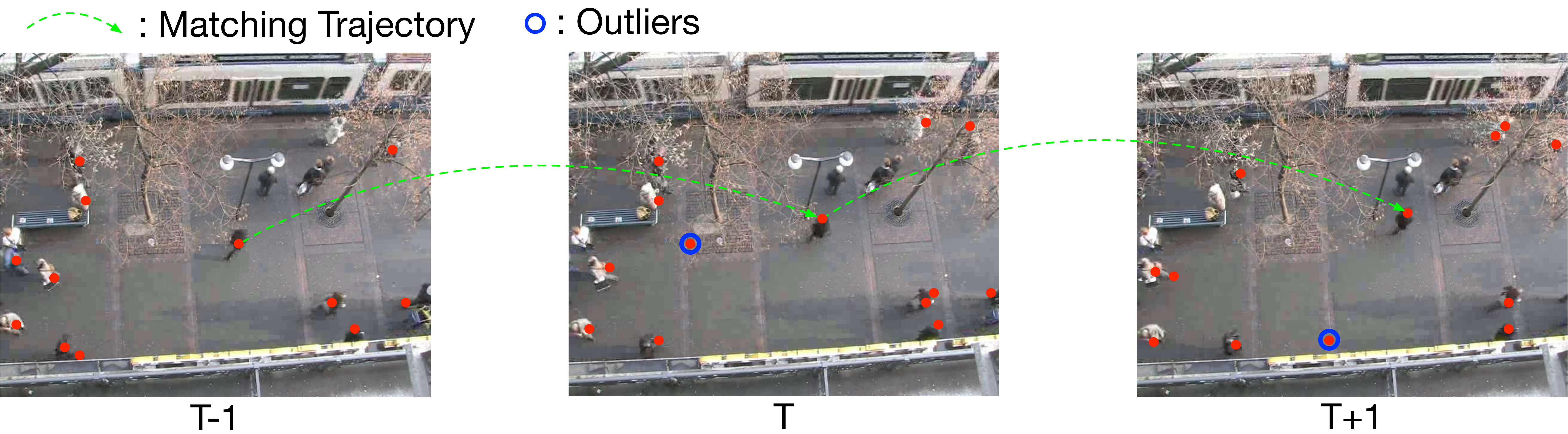}
\end{center}
\vspace{-4ex}
   \caption{An example of instance matching. Green dashed line denotes the instance points matching across timesteps. Blue circles denote the outliers of the instance points (also mean missmatching points).}
\label{fig:km}
\vspace{-3ex}
\end{figure*}

The instance points obtained from the clustering method are independent across time. To obtain the trajectory, we perform cross-time instance matching. The basic idea is to establish a cost matrix between two consecutive images where each entry indicates the distance between two instance points across two images. Then we apply the Kuhn-Munkres (KM) algorithm~\footnote{\url{http://software.clapper.org/munkres/}} to calculate the minimum-cost matching. To better incorporate the appearance feature, we also use the RGB information as a part of distance. The final distance function is designed as
$\mathcal{D}_{ij} = dist(P_i, Q_j) + \lambda rgb(P_i,Q_j)$, where $P_i$ and $Q_j$ are two instance points from two images. $dist(\cdot)$ is the Euclidean distance and $rgb(\cdot)$ is the L1 distance.

Specifically, the cost matrix $\mathcal{C}\in \mathbb{R}^{M\times N}$ is defined as the all-to-all distance between two images, where $M$ and $N$ indicate the number of instance points in two images. We use the KM algorithm with the cost matrix $\mathcal{C}$ to get the minimum-cost matching. The workflow is shown in Fig.~\ref{fig:km}. The matching pair from the KM algorithm is specified as a \emph{true} pair if its distance is less than the pre-defined threshold $\mathbb{D}$, otherwise it is a \emph{false} pair.
Note that there exist some outliers that do not match any point. We label these outliers with blue circles. To handle these points, we apply some specific methods to filter them. For the points in image $T$, if we cannot find the former matching points in image $T-1$ but can find the matching points in image $T+1$, we label these points as the starting points of the sequence, otherwise we label them as outliers. This bidirectional filter benefits the precision of cross-time matching.

\subsection{Trajectory Prediction}
After extracting the trajectories in the pixel coordinate system, we transfer them to the real-word coordinate system and use them as the dataset to train and validate the prediction network in the last stage. At any time $t$, the status for the $i$th dynamic instance can be represented as the location $p_i=(x_i^t, y_i^t)$. The task for the prediction network is to observe the status of all the dynamic instances in the time interval $[1 : T_{obs}]$ and then predict their discrete positions at $[T_{obs}+1 : T_{pred}]$. We have highlighted many learning-based works in Section 2.1 and these methods can be directly used in our approach. Because the datasets we use are human crowd videos, we utilize some classical LSTM-based approaches for pedestrian trajectory prediction in our experiments to verify the effectiveness of our unsupervised method.

\subsection{Optimization}
In the proposed approach, dynamic-point modeling and trajectory prediction stages have trainable parameters, and the other two stages are non-parametric. The whole workflow is stage-by-stage.
We first train the dynamic-point modeling part. An ADAM optimizer with learning rate = 1e-4 is used for optimization. $\beta$ is 0.5, $\sigma$ is 0.1 and $\lambda$ = 0.2.
Then we apply the well-trained dynamic-point extractor to access the dynamic points. After dynamic-to-instance and instance matching, we get the extracted trajectories. 
For the trajectory prediction part, we follow the settings in the original paper to train the network optimizer, including the observation and prediction length.

\subsection{Network Architecture}
\noindent \textbf{Dynamic-Point Modeling.} For the dynamic-point extractor, we use the basic block (Conv2d + BatchNorm2d +Leaky Relu) in VGG~\cite{vgg} as the unit. The sizes of Conv2d are: [64, 128, `M', 256, 256, `M', 512, 512, `M', 512, 512], where `M' denotes the MaxPooling and each number indicates the size of one unit.
For the encoder, we use a structure similar to the dynamic-point extractor.
For the decoder part, we use the reverse setting of the encoder to keep the output and input size consistent. The detailed setting is [512, 512, `U', 256, 256, `U', 256, 256, `U', 128, 64], where `U' denotes the bilinear upsampling.

\section{Experiments}

\subsection{Implementation Details}
For trajectory prediction, we use several LSTM-based models, including Vanilla-LSTM, Occupancy-LSTM (O-LSTM), and Social LSTM (S-LSTM)~\cite{Alahi2016SocialLH}. They are trained by ground truth data before. In our approach, we use our extracted trajectories to train these models. Following the original setting in S-LSTM, we filter our extracted trajectories by removing the trajectories with lengths less than 20 frames (8 seconds). We set $K$=180 so that the dynamic points could distribute all moving objects.

\noindent \textbf{Evaluation Metrics.} We evaluate our performance on three aspects: detected instance points, extracted trajectories, and predicted trajectories.

We introduce recall and precision to test the quality of instance points extracted from Dynamic-to-Instance Aggregation.
We give the detailed explanation as follows. (1) True-Positive instance points: instance points where the distance between detected instance points and the ground-truth points is less than the threshold $\mathcal{D}$. 
(2) Recall: the ratio of True-Positive points to all ground-truth points.
(3) Precision: the ratio of True-Positive points to all detected instance points. We term them Ins-Recall and Ins-Precision, respectively.

We also apply recall and precision to test qualities of extracted trajectories. The True-Positive trajectories are defined as: trajectories where the average distance between extracted trajectories and ground truth trajectories across timesteps is less than the threshold $\mathcal{E}$. The definition of recall and precision is similar to the statement above. We term them Gen-Recall and Gen-Precision, respectively.
Note that there exist some conditions where one detected instance point (or trajectory) corresponds to several ground truth points (or trajectories), or vice versa. We use the KM algorithm to get the minimum cost matching. Both precision and recall are calculated on average. We set $\mathcal{E}$ = 1.5 and $\mathcal{D}$ = 1.5. 

Similar to prior work~\cite{Alahi2016SocialLH}, we use two popular evaluation metrics for predicted trajectory evaluation: 
(1) Average Displacement Error (ADE): Average L2 distance between predicted trajectory and the ground truth over all timesteps.
(2) Final Displacement Error (FDE): The distance between the predicted final destination and the true final destination in the ground truth. Besides the comparison between our unsupervised method and supervised methods, we also conduct semi-supervised experiments by using our extracted trajectories as extra data to train supervised models.

\subsection{Datasets}
For the dynamic-point modeling part, we use two publicly available datasets: ETH~\cite{ETH} and UCY~\cite{UCY} as the training data. These two datasets are captured by fixed-cameras. Although there are some other datasets containing videos of traffic scenarios such as KITTI~\cite{Geiger2013VisionMR} and Argoverse~\cite{Chang2019Argoverse3T}, the videos are all captured in drivers' view. The camera is moving and and they do not provide the homograph matrix for each frame, which is not infeasible for our method.

We follow Social LSTM~\cite{Alahi2016SocialLH} to split the video to frames at every 0.4 seconds. For the trajectory prediction stage, we need to convert pixel coordinates to real-world coordinates to train these LSTM-based methods.
Therefore, the extrinsic matrix is required to transfer the pixel coordinates to the real-world coordinates. From the open-source codebase~\footnote{\url{https://github.com/trungmanhhuynh/Scene-LSTM}}, it can be found that only three scenes (UCY-Zara01, UCY-Zara02, and UCY-University) have complete transform matrixes. We thus use these three scenes for trajectory prediction. 

\begin{table}
\caption{Evaluation results of detected instance points. We compare the proposed method with the unsupervised tracking~\cite{He2018TrackingBA} method and unsupervised keypoint modeling method~\cite{Minderer2019UnsupervisedLO}. `-' indicates the model cannot converge in the dataset}
\small
\setlength{\tabcolsep}{0.2pt}
\centering
\begin{tabular*}{1.0\linewidth}{c|c|c|c|c|c|c|c|c|c|c}
\toprule
Metric  & \multicolumn{5}{c|}{Ins-Precision} & \multicolumn{5}{c}{Ins-Recall} \\
\hline
Dataset & ETH & Hotel & Univ & Zara1 & Zara2 & ETH & Hotel & Univ & Zara1 & Zara2 \\
\hline
Un-Tracking~\cite{He2018TrackingBA} & 8.3\% & - & - & 19.6\% & 21.4\% & 12.7\% & - & - & 10.1\% & 14.8\% \\
\hline
Un-Keypoint~\cite{Minderer2019UnsupervisedLO} & 16.8\% & 11.2\% & - & 33.1\% & 36.7\% & 14.1\% & 14.6\% & - & 39.4\% & 41.0\% \\
\hline
Ours & \textbf{47.9}\% & \textbf{37.1}\% & \textbf{36.4}\% & \textbf{58.7}\% & \textbf{60.3}\% & \textbf{58.3}\%  & \textbf{42.0}\% & \textbf{31.4}\% & \textbf{63.1}\% & \textbf{67.9}\% \\

\bottomrule
\end{tabular*}
\label{tab:ip1}
\vspace{-4ex}
\end{table}

\subsection{Results}

\subsubsection{Experimental Results for Instance Points.}We first evaluate the extracted instance points on various datasets. Since there is no annotation in any of the datasets, we use the unsupervised object tracking algorithm~\cite{He2018TrackingBA} and the keypoint-based video prediction algorithm~\cite{Minderer2019UnsupervisedLO} as baseline methods. 
From Table~\ref{tab:ip1}, several phenomena can be found: 1) in all datasets, the proposed dynamic-point modeling and dynamic-to-instance aggregation achieve consistently better performance than unsupervised tracking and unsupervised keypoint modeling; 2) for Hotel and Univ (where there are a large number of moving instances), unsupervised tracking method cannot converge while our method remains generalizable; 3) unsupervised keypoint modeling method without considering the sequential temporal information also performs poor (even does not converge in Univ dataset), while our method exploits the temporal consistency and achieves decent performance for all videos. Hence, for unsupervised tracking and keypoing modeling methods, it is difficult for them to extract dynamic instances from these videos, which are in bird's-eye view containing limited appearance and structure features. 
Instead, the proposed dynamic-point modeling and dynamic-to-instance aggregation could better handle the difficulties. 

\begin{figure*}[t]
\begin{center}
   \includegraphics[width=1.0\linewidth]{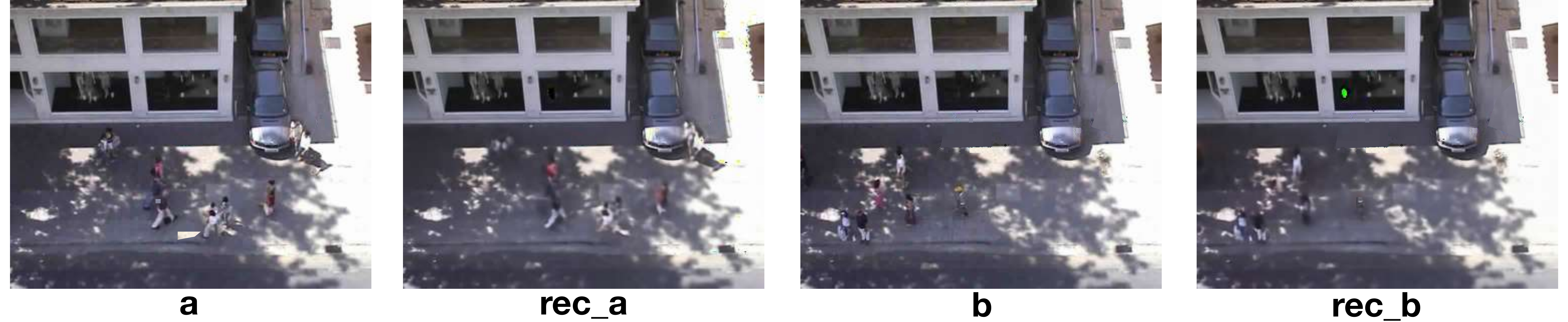}
\end{center}
\vspace{-3ex}
   \caption{The input image \emph{vs.} the reconstructed image from the decoder. \textbf{a} and \textbf{b} are the input images, and \textbf{rec\_a} and \textbf{rec\_b} are the reconstructed images.}
\label{fig:recon}
\end{figure*}

\noindent \textbf{Visualization for the decoder.} To investigate the performance of the decoder part in dynamic-point modeling, we visualize the reconstructed images in Fig.~\ref{fig:recon}. It can be observed that the moving pedestrians are well captured and reconstructed, even with a large number of moving objects. The reconstructed images are also real and decent. Hence, it can be found that our dynamic modeling does capture the dynamic information and could reconstruct the input image.

\begin{figure*}[t]
\begin{center}
   \includegraphics[width=1.0\linewidth]{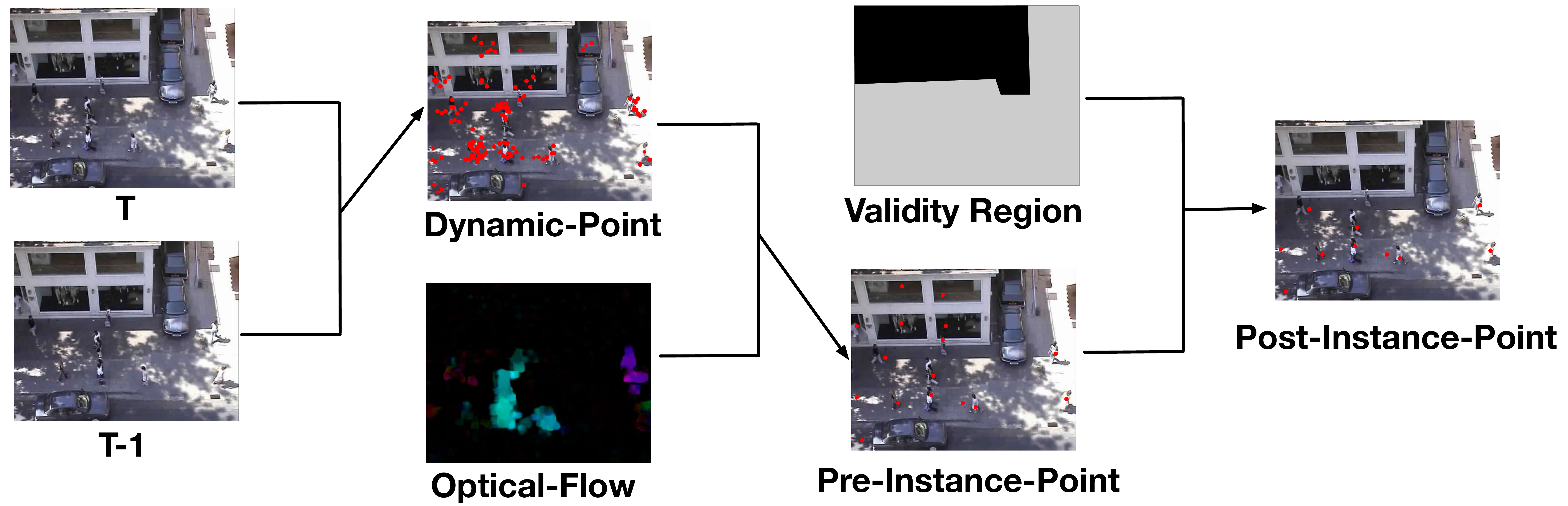}
\end{center}
\vspace{-3ex}
   \caption{Visualization for the output of each step in dynamic-to-instance aggregation. For the image of the valid region, the grey color denotes the valid part while the black color indicates the invalid region.}
\label{fig:ip1}
\vspace{-3ex}
\end{figure*}

\noindent \textbf{Visualization for each step in dynamic-to-instance aggregation.} To give a more intuitive description, we visualize the output of each step during instance-point extraction in Fig.~\ref{fig:ip1}. Specifically, we first use Image (\textbf{T}) and Image (\textbf{T-1}) to extract the dynamic points. Then both dynamic points and optical flow are used to get the pre-instance points. Due to some invalid regions (buildings, railways) for pedestrians, we constrain these instance points with the valid region map. Because the background for a fixed-camera is static, it is easy to circle the valid region on just one frame. After that, we obtain the post-instance points.

\subsubsection{Experimental Results for extracted trajectories.} 
\begin{figure*}[t]
\begin{center}
   \includegraphics[width=1.0\linewidth]{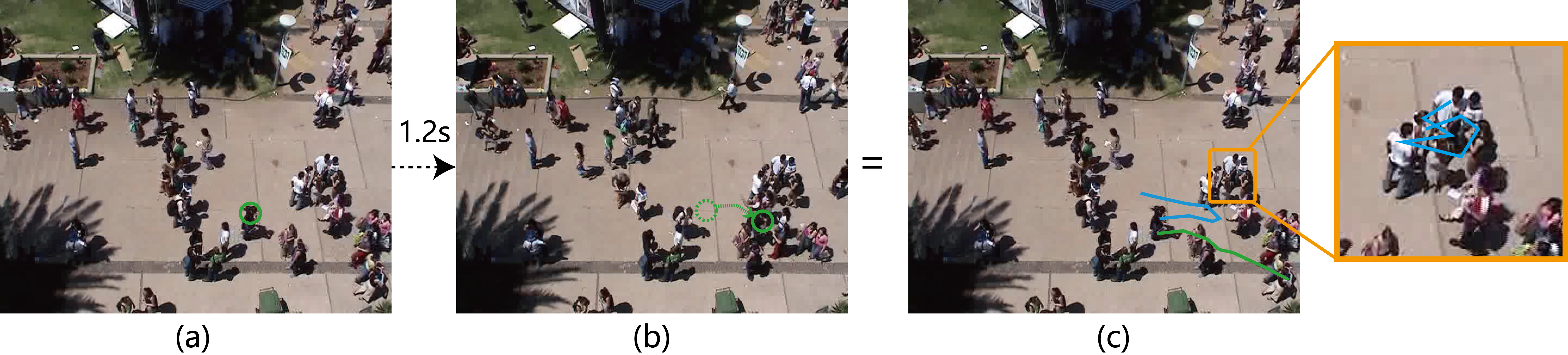}
\end{center}
\vspace{-3ex}
   \caption{Bad cases of trajectory extraction. We label a pedestrian in Image (a) with a solid green circle. Image (b) is the scene from (a) after 1.2 seconds (the dashed line denotes the trajectory during the 1.2 seconds). We show GT and ET in Image (c). We zoom in a bad case in the yellow box that when a group of people stand around talking and they still have slight movement, the extracted trajectory would be in mess.}
\label{fig:bascase1}
\vspace{-2ex}
\end{figure*}

We use Gen-Recall and Gen-Precision to test the performance of extracted trajectories. Three datasets, including Zara1, Zara2, and Univ, are reported. The results are shown in the following; Gen-Precision of Zara1, Zara2 and Univ is 49.1\%, 53.7\%, and 23.7\% respectively. Gen-Recall of Zara1, Zara2 and Univ is 52.9\%, 54.4\%, and 20.6\% respectively. Our method could generate about a half number of trajectories similar to the ground truth for general videos. We visualize some extracted trajectories in Fig.~\ref{fig:results}. For very crowded scene (Univ), the performance drops due to the mismatching of instances. We show some bad cases in Fig.~\ref{fig:bascase1}. When multiple pedestrians meet, the error of instance matching occurrs and the trajectories of these pedestrians are biased in the wrong direction. It is also a fundamental obstacle for multi-object tracking methods.

\begin{table}
\caption{Experimental results of trajectory prediction. We use three popular models to test the extracted trajectories, where O-LSTM and S-LSTM are both from Social-LSTM~\cite{Alahi2016SocialLH}. LSTM(sup), O-LSTM(sup), and S-LSTM(sup) indicate these models in a supervised manner. The unit for ADE and FDE is meters}
\small
\setlength{\tabcolsep}{7pt}
\centering
\begin{tabular*}{0.85\linewidth}{c|c|c|c|c|c|c}
\toprule
Metric & \multicolumn{3}{c|}{ADE} & \multicolumn{3}{c}{FDE}  \\
\hline
Dataset & Univ & Zara1 & Zara2 & Univ & Zara1 & Zara2 \\
\hline
LSTM & 0.936 & 0.729 & 0.742 & 1.512 & 1.24 & 1.338  \\
\hline
O-LSTM & 0.875 & 0.511 & 0.579 & 1.427 & 0.947 & 1.092  \\
\hline
S-LSTM & 0.892 & 0.477 & 0.495 & 1.45 & 0.911 & 1.03  \\
\hline
LSTM (sup) & 0.52 & 0.43 & 0.52 & 1.25 & 0.93 & 1.09  \\
\hline
O-LSTM (sup) & 0.35 & 0.22 & 0.28 & 0.90 & 0.46 & 0.58  \\
\hline
S-LSTM (sup) & 0.27 & 0.22 & 0.25 & 0.77 & 0.48 & 0.50  \\
\bottomrule
\end{tabular*}
\label{tab:tp2}
\vspace{-2ex}
\end{table}

\subsubsection{Experimental Results for Trajectory Prediction.}
\begin{figure}[htb]
\subfigure[]{
\label{fig:result_2}
\includegraphics[width=0.32\columnwidth]{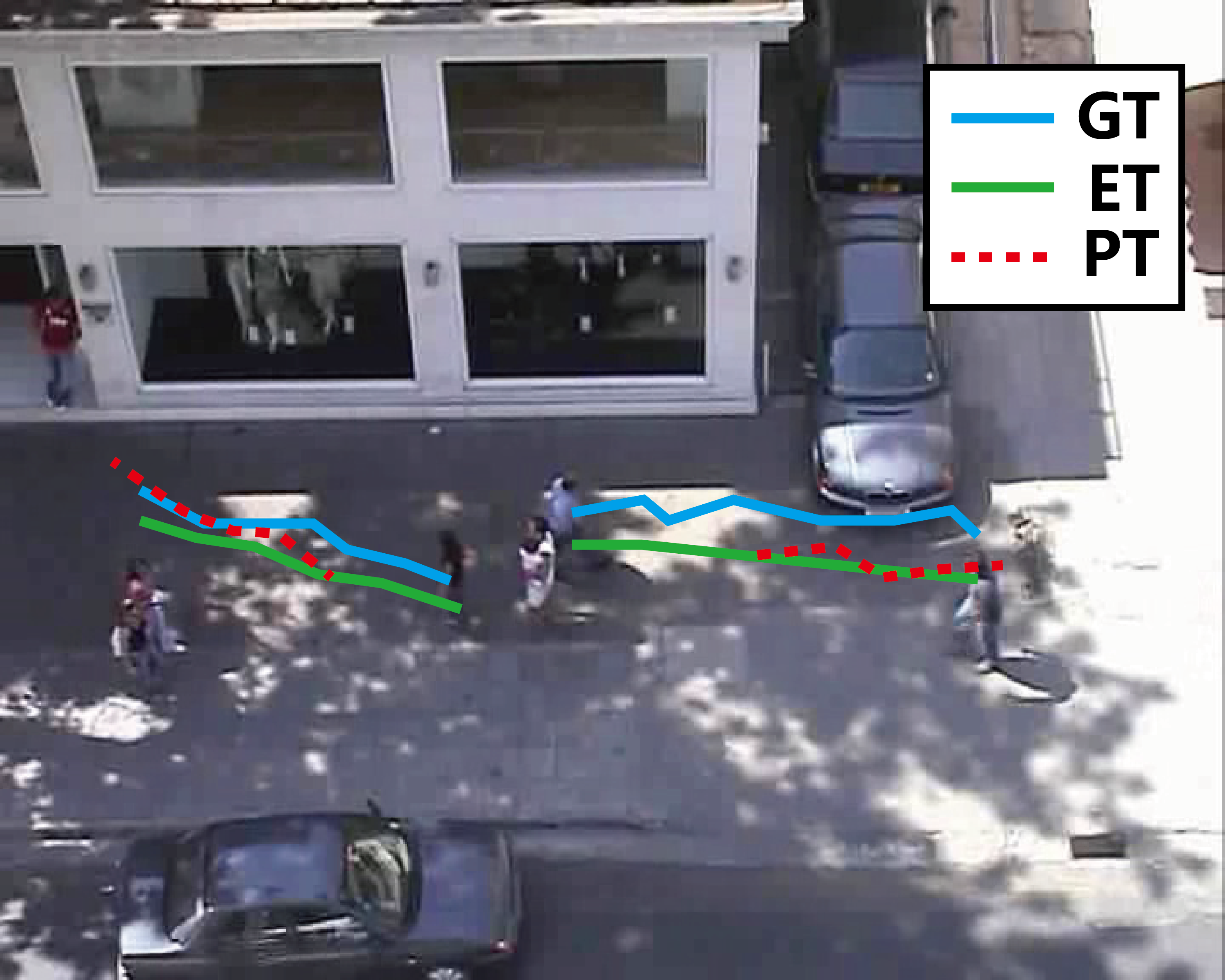}}
\subfigure[]{
\label{fig:result_3}
\includegraphics[width=0.32\columnwidth]{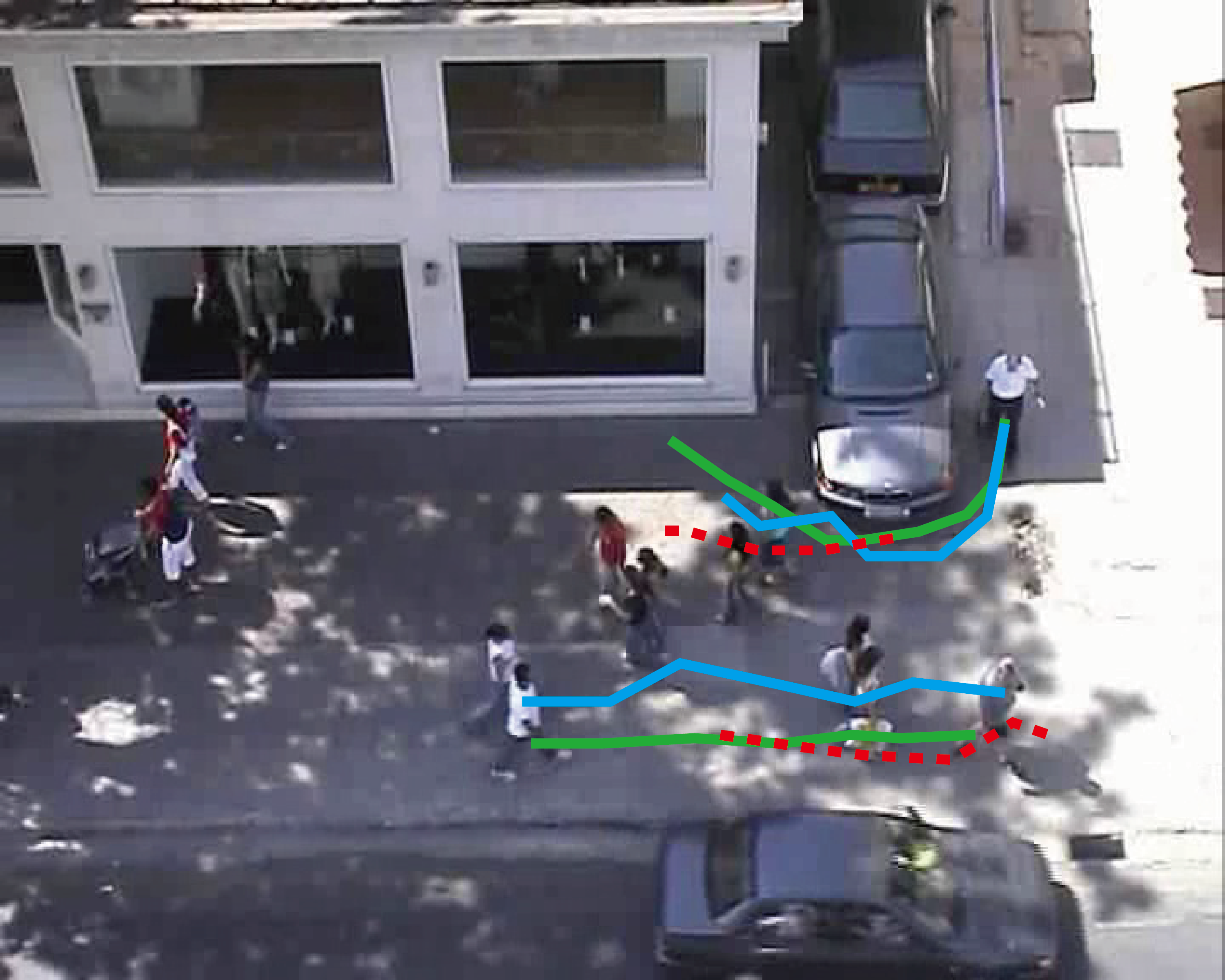}}
\subfigure[]{
\label{fig:result_4}
\includegraphics[width=0.32\columnwidth]{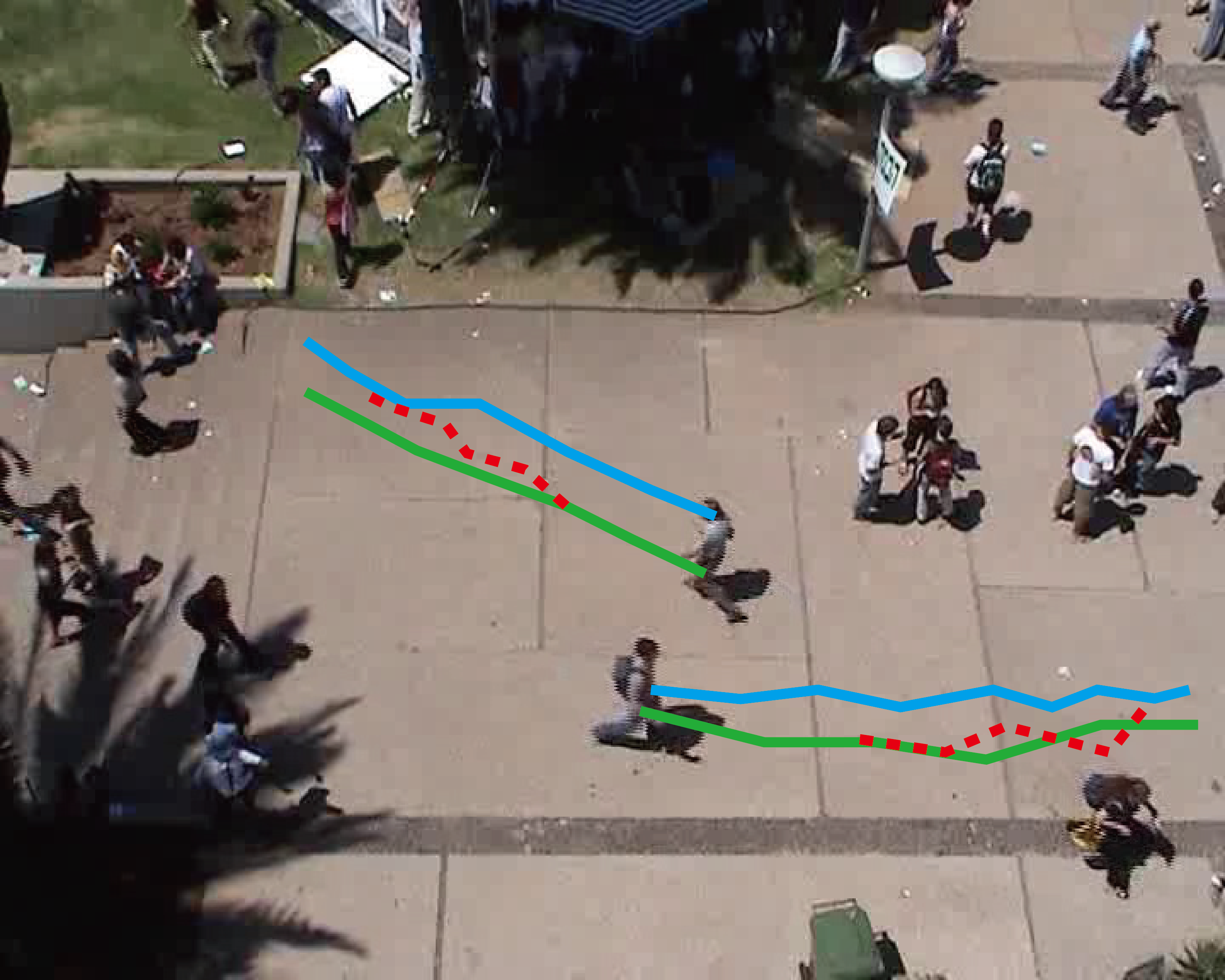}}
\vspace{-2ex}
\caption{Visualization for trajectory prediction. We display three examples with the ground truth trajectory (GT in green line), the extracted trajectory by our method (ET in blue line), and the predicted trajectory by our method (PT in red dashed line).}
\label{fig:results}
\vspace{-2ex}
\end{figure}

Because the datasets (Zara1, Zara2, and Univ) we use are about pedestrians, we test the extracted trajectories with three popular models for predicting trajectories of pedestrians, including LSTM, O-LSTM, and S-LSTM. We use a popular evaluation method, the leave-one-out approach, to test the trajectory prediction part, where we train on 2 scenes and test on the remaining one. We follow settings from prior works to observe the trajectory for 8 timesteps (3.2s) and predict the trajectory of 12 timesteps (4.8s). We use our extracted trajectories in the training process and test with the ground truth. The results of trajectory prediction are shown in Table.~\ref{tab:tp2}. The performance on Univ is worse than the other two scenes because it is more complex with a crowd of moving objects. We also display the performances of LSTM, O-LSTM, and S-LSTM with supervision. We can see that the supervised method performs better than our unsupervised methods. It is mainly because our extracted trajectories are not smooth (Fig.~\ref{fig:results}) as the ground truth and sometimes we have bad cases (Fig.~\ref{fig:bascase1}). However, for our unsupervised method without any label, the ADE is about half meter and FDE is about one meter, it still has good practical significance.


\noindent \textbf{Visualization for trajectory prediction.} In Fig.~\ref{fig:results}, We show several examples to display the ground-truth trajectory, our extracted trajectory, and our predicted trajectory. From the visualization, we can find that the extracted trajectories mainly focus on the centers of moving objects, which demonstrates that our generated instance points can capture the main dynamic information of moving objects. After training on extracted trajectories, our trajectory predictor can also work on true trajectories, which also illustrates the usefulness of our extracted trajectories in an unsupervised manner.

\noindent \textbf{Semi-supervised training for trajectory prediction.} To show the capability of our extracted trajectories in improving current supervised prediction models, we conduct semi-supervised experiments. We first use the ground truth data of Zara1 to train the model. Then we use extracted trajectories from other datasets as extra data to further train the model. Table~\ref{tab:semi} shows the results of testing on Zara2. We can see that adding more our extracted trajectories in the training process will make the prediction results more accurate. It illustrates our method is feasible in using large-scale raw videos to improve current models.

\begin{table}
\vspace{-2ex}
\caption{Results of Semi-supervised training }
\small
\setlength{\tabcolsep}{4.5pt}
\centering
\begin{tabular*}{0.88\linewidth}{c|c|c|c|c|c|c}
\toprule
Dataset  & \multicolumn{2}{c|}{Zara1} & \multicolumn{2}{c|}{+Univ(Gen)} & \multicolumn{2}{c}{+Univ(Gen)+Zara2(Gen)} \\
\hline
Method &  LSTM & S-LSTM &  LSTM & S-LSTM & LSTM & S-LSTM  \\
\hline
ADE & 0.598 & 0.347   & 0.578 & 0.341 & 0.521 & 0.320  \\
\hline
FDE & {1.25} & {0.69}   & {1.157} & {0.687} & 1.094 & 0.659  \\
\bottomrule
\end{tabular*}
\label{tab:semi}
\vspace{-6ex}
\end{table}

\subsection{Ablation Study}
In this section, we perform several ablation studies to investigate the effectiveness of different components of the proposed approach. We train the dynamic-point modeling part with all five scenes and test the performance on Zara1 and Zara2.

\noindent \textbf{Components of Clustering.} For the dynamic-to-instance aggregation part, we use two types of dynamic information as the features, \ie~gaussian heatmaps and optical flow. From Table~\ref{tab:ab1}, it can be found after removing the dynamic points and optical flow, the performance of instance points is about 20\% worse. Additionally, the model without dynamic points performs worse than the model without optical flow, which also demonstrates that dynamic points play a more important role in the instance-point extraction.

\noindent \textbf{Forward \emph{vs.} Backward Extractors.}
In the dynamic-point modeling part, we apply a forward-backward cycle extractor to keep the dynamic points consistent in cycle timesteps. We try to remove one of them to perform the ablation study. From Table~\ref{tab:ab1}, it can be observed that removing the forward extractor or removing the backward extractor will decrease the performance. Both forward and backward extractors are important ingredients in the instance-point extraction. 

\noindent \textbf{Consistency Loss.} Moreover, we remove the consistency loss between the forward and backward extractors to check the effect. The results in Table~\ref{tab:ab1} show that the consistency loss further boosts the forward-backward extractors (about 3\%-4\%) during the instance-point extraction.

\noindent \textbf{Scalability.} To verify the scalability of the proposed dynamic-point modeling, we compare the model trained with all five scenes to the model trained with only two scenes (Zara1 and Zara2). The results in Table~\ref{tab:ab1} show that more video data improves the performance. It also demonstrates that our methods keep good scalability and take full advantage of large-scale video data.

\begin{table}
\vspace{-2ex}
\caption{Ablation studies for instance-point extraction. We make several variants to investigate the effectiveness of different components}
\small
\setlength{\tabcolsep}{6.5pt}
\centering
\begin{tabular*}{0.85\linewidth}{c|c|c|c|c}
\toprule
Metric  & \multicolumn{2}{c|}{Ins-Precision} & \multicolumn{2}{c}{Ins-Recall} \\
\midrule
Dataset &  Zara1 & Zara2 & Zara1 & Zara2 \\
\midrule
Ours w/o Dynamic-Point &  38.3\% & 39.8\% & 44.1\% & 48.2\% \\
\hline
Ours w/o Optical Flow & 40.7\% & 43.4\%   & 49.9\% & 53.1\%  \\
\hline
\hline
Ours w/o Forward Extractor & 46.8\% & 50.1\%   & 54.4\% & 59.8\%  \\
\hline
Ours w/o Backward Extractor & 52.1\% & 56.8\%   & 57.8\% & 62.2\%  \\
\hline
\hline
Ours w/o Consistency Loss& 56.2\% & 58.0\%   & 59.1\% & 60.4\%  \\
\hline
\hline
Ours w/ only-two-scenes & 52.3\% & 53.9\%   & 58.1\% & 61.6\%  \\
\hline
\hline
Ours & \textbf{58.7}\% & \textbf{60.3}\%   & \textbf{63.1}\% & \textbf{67.9}\%  \\
\bottomrule
\end{tabular*}
\label{tab:ab1}
\vspace{-6ex}
\end{table}

\subsection{Limitations and Future Work}

Although the proposed method works in an unsupervised manner, there also exist some limitations. 1) The whole framework is not end-to-end. We train these learnable components one by one. 2) There are some hyper-parameters, which need fine-tuning when training with different datasets. 3) Since there is no target category, our method might focus on the dynamic part of non-target category, such as a car in the pedestrian trajectory dataset. We visualize some badcases in the supplementary materials.
In the future work, we aim to incorporate the category-aware memory and template into the dynamic modeling to further distinguish different categories. And we will also explore dynamic point-based approach on drivers' view videos.

\section{Conclusion}
In this paper, we propose a complete pipeline for label-free trajectory extraction and prediction. To our knowledge, this is the first time unsupervised trajectory extraction and prediction have been explored. We make full use of the spatial consistency by image reconstruction and the temporal dynamic consistency by sequential frames to capture moving regions in videos through dynamic points. To extract trajectories at the instance-level, we also propose a novel aggregation approach to cluster dynamic points to instance points by compensating with optical flow. Without any supervision, our method uses raw videos to extract trajectories and train trajectory prediction networks. The experiments show the effectiveness and scalability of our approach.  

\clearpage
%
%
\bibliographystyle{splncs04}
\bibliography{egbib}
\end{document}